\documentclass{article}
\usepackage{amsmath}  

\usepackage{PRIMEarxiv}

\usepackage[utf8]{inputenc} 
\usepackage[T1]{fontenc}    
\usepackage{hyperref}       
\usepackage{url}            
\usepackage{booktabs}       
\usepackage{amsfonts}       
\usepackage{nicefrac}       
\usepackage{microtype}      
\usepackage{lipsum}
\usepackage{fancyhdr}       
\usepackage{graphicx}       
\graphicspath{{media/}}     
\usepackage{tcolorbox} 
\usepackage{float} 

\title{Reinforcement Learning is all you need
}

\author{
  Yongsheng Lian \\
  Mechanical Engineering Department \\
  University of Louisville \\
  Louisville, KY 40223\\
  \texttt{yongsheng.lian@louisville.edu} \\
}

\begin{document}
\maketitle

\begin{abstract}

Inspired by DeepSeek R1’s success in reasoning via reinforcement learning without human feedback, we train a 3B language model using the Countdown Game with pure reinforcement learning. Our model outperforms baselines on four of five benchmarks, demonstrating improved generalization beyond its training data. Notably, response length does not correlate with reasoning quality, and while "aha moments" emerge, they do not always yield correct answers. These findings highlight the potential of RL-only training for reasoning enhancement and suggest future work on refining reward structures to bridge emergent insights with accuracy.

\end{abstract}


\section{Introduction}

Post-training plays a crucial role in refining language models, ensuring they exhibit good reasoning capabilities, alignment with ethical and social values, and adaptability to user-specific preferences. Unlike pre-training, which demands extensive computational resources and large-scale datasets, post-training is a more efficient process that leverages targeted fine-tuning techniques such as reinforcement learning from human feedback (RLHF) \cite{christiano2023deepreinforcementlearninghuman, ouyang2022traininglanguagemodelsfollow}, instruction tuning \cite{wei2022finetunedlanguagemodelszeroshot}, and Direct Preference Optimization (DPO) \cite{rafailov2024directpreferenceoptimizationlanguage}. These approaches enable models to generalize better across various tasks and mitigate biases. Furthermore, post-training requires significantly low computational overhead compared to pre-training \cite{guo2025deepseek}.

Reasoning has long been regarded as a cornerstone in the development of large language models (LLMs). To effectively serve diverse purposes, LLMs must demonstrate a substantial level of reasoning proficiency. In the early stages of enhancing reasoning in language models, researchers focused on methods like prompt engineering, which involved carefully designing input prompts to guide the model's thought process and elicit more accurate and logical responses. One early breakthrough was the use of Chain-of-Thought (CoT) reasoning, which encouraged models to generate intermediate steps while solving complex tasks, allowing for improved reasoning in areas such as mathematics and logical problem-solving \cite{wei2023chainofthoughtpromptingelicitsreasoning}.

As the field advanced, OpenAI’s O-series models introduced more sophisticated techniques, such as inference-time scaling, which expanded the CoT process to longer and more intricate reasoning chains. This approach resulted in substantial improvements in tasks requiring advanced reasoning~\cite{OpenAI2024b}. Other approaches, such as the Monte Carlo Tree Search (MCTS) \cite{feng2023alphazero, gao2024interpretable} and process-based reward models (PBM) \cite{lightman2023letsverifystepstep, wang2023math} have also been used to enhance reasoning capabilities.

\begin{figure}
  \centering
  \includegraphics[width=0.8\textwidth]{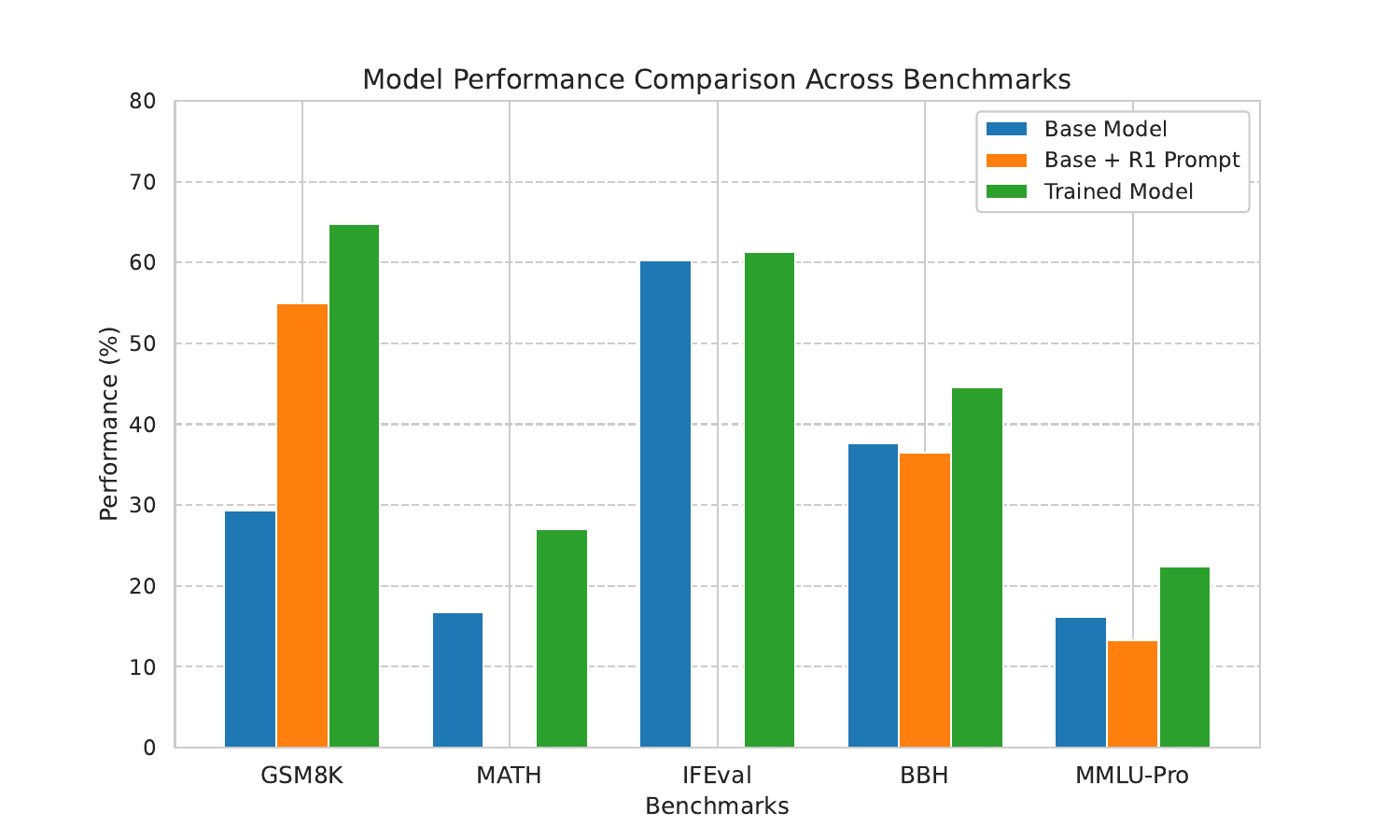}  
  \caption{Model performance on different benchmarks}
  \label{fig:completion_length}
\end{figure}

In reinforcement learning, Supervised Fine-Tuning (SFT) is commonly employed to adapt pre-trained models to specific tasks by utilizing labeled data. This approach enables models to learn task-specific features and patterns, thereby enhancing their accuracy and relevance. However, the reliance on supervised data collection can be time-consuming and resource-intensive.

Reinforcement learning (RL) has shown significant potential in enhancing reasoning capabilities. One example is the AlphaGo where RL was used to teach the model to play the game of Go at a superhuman level. They  trained  deep neural networks by a combination of both supersived learning from human expert games and reinforcement learning from games of self-play~\cite{silver2016mastering}. Another example is DeepSeek R1~\cite{shao2024deepseekmathpushinglimitsmathematical, guo2025deepseek}. The model developed improved reasoning capabilities through rule-based reinforcement learning. They introduced a training pipeline that has two RL stages  and two SFT stages.  

The first SFT stage (Cold Start) begins by collecting thousands of long CoT examples as cold-start data. The DeepSeek-V3-Base model is then fine-tuned on this cold-start data. This initial SFT aims to provide the model with a foundation for reasoning. The first RL stage following the cold-start SFT is a large-scale reinforcement learning process. This stage focuses on enhancing the model's reasoning capabilities. During this RL, a language consistency reward is introduced to mitigate language mixing in the CoT. 

The second SFT stage starts once the reasoning-oriented RL converges. The resulting checkpoint is used to collect new SFT data through rejection sampling. This new SFT data is more diverse than the initial cold-start data. It uses about 600k reasoning-related training samples and around 200k training samples unrelated to reasoning. The second SFT stage aims to enhance the model's capabilities in both reasoning and general-purpose tasks. 

After the second SFT stage, the checkpoint undergoes an additional RL process. This final RL stage aims to further align the model with human preferences, improving its helpfulness and harmlessness while simultaneously refining its reasoning capabilities. This stage uses a combination of reward signals and diverse prompt distributions, taking into account prompts from all scenarios (reasoning and general data). While SFT is used in training, RL is  the primary driver for advanced reasoning.

In this paper, we demonstrate that reinforcement learning can enhance the reasoning capabilities of language models. Our trained model outperforms the base model on four out of five benchmarks, showcasing the effectiveness of RL in improving generalization beyond the training data. We also replicate the phenomenon of the "aha moment" as reported in previous work \cite{guo2025deepseek}, where the model exhibits moments of sudden insight.  We find that the quality of reasoning decreases with longer responses in our experiments, highlighting a complex relationship between response length and reasoning quality.

\section{Method}

\subsection{Training Dataset}

To train our language models, we utilize the Countdown Game, a well-known numerical puzzle that challenges players to reach a given target number using a randomly drawn set of integers and basic arithmetic operations ($+, -, \times , \div $) \cite{tinyzero}. For instance, if the available numbers are 6, 7, 8, and 9, and the target number is 24, a valid solution would be $ 8 \times (6 \div (9 - 7)) = 24 $. This game serves as an ideal dataset for training models in numerical reasoning, as it provides a structured yet diverse set of problem-solving scenarios. Furthermore, the correctness of solutions is straightforward to verify, making it a reliable benchmark for evaluating model performance in arithmetic problem-solving.

\subsection{Rule-Based Reward Modeling}

To guide model behavior effectively, we employ a simple yet structured rule-based reward model that evaluates both the response format and the correctness of the generated solution. Specifically, we define two distinct reward mechanisms: the Format Reward and the Answer Reward. The prompts used for generating responses are adapted from \cite{guo2025deepseek} and are structured as follows:  

\paragraph{Format Reward}

\begin{itemize}
\item \textbf{system}: You are a helpful assistant. You first think about the reasoning process in the mind and then provide the user with the answer.
\item \textbf{user}: f"Using the numbers {numbers}, create an equation that equals {target}. You can use basic arithmetic operations (+, -, *, /) and each number can only be used once. Show your work in <think> </think> tags. And return the final equation and answer in <answer> </answer> tags. For example, <answer> (1 + 2) / 3 = 1 </answer>."
\item \textbf{assistant}: Let me solve this step by step.<think>
\end{itemize}

The Format Reward ensures structural compliance by enforcing a well-formed response. Using regular expression-based extraction, we verify that the response begins with a <think> tag, ensuring that reasoning is explicitly articulated before the final answer.  The <think> tag significantly reduces the difficulty for the base model to follow instructions \cite{xie2025logicrlunleashingllmreasoning}. It also ensure the response does not contain nested <think> tags. The \textit{<answer>...</answer>} section ensures that a final equation and result are clearly presented after the reasoning process.  This reward mechanism serves as a fundamental safeguard, ensuring that the model adheres to the expected reasoning structure before evaluation of correctness.  

\paragraph{Answer Reward}

The Answer Reward evaluates the correctness of the final result computed by the model. We use a simple binary answer reward function. If the answer within the <answer> </answer> tag matches the target number, the reward model receives a reward of 1; otherwise, the reward is 0. While this approach offers a straightforward correctness check, it does not consider partial progress or important intermediate steps. A more advanced reward structure such as Process Reward Models (PRMs)~\cite{lightman2023letsverifystepstep, zhang2025lessonsdevelopingprocessreward} would assign partial credit for logically sound reasoning, even if the final answer is incorrect. The PRMs can promote structured problem-solving and reduces the risk of penalizing models for minor calculation errors that do not invalidate the overall logical process.

By integrating both format validation and answer correctness, our reward model offers an effective mechanism for guiding the model to generate coherent, accurate, and logically consistent numerical solutions.

\subsection{Reinforcement Learning Algorithm}
\label{sec:GRPO}

The DeepSeek-R1 training used the Group Relative Policy Optimization (GRPO) method\cite{guo2025deepseek}. It is an alternative to tranditional reinforcement learning methods such as the Proximal Policy Optimization (PPO)\cite{schulman2017proximalpolicyoptimizationalgorithms}. The key distinction between GRPO and PPO lies in the handling of the value function. In PPO, the value function serves as a baseline for computing the advantage function, typically learned via a neural network. In contrast, GRPO eliminates the need for an explicit value function, reducing computational complexity while still maintaining a relative measure of reward effectiveness.

Instead of relying on a learned value function, GRPO generates a set of responses $\{o_1, o_2, \ldots, o_G\}$ from the old policy $\pi_{\theta_{old}}$  and then optimizes the policy model $\pi_\theta$ and uses their rewards to calculate each response’s advantage as below:


\begin{equation}
  \begin{aligned}
    \mathcal{J}_{GRPO}(\theta) = \mathbb{E}\Bigg[ \sum_{i = 1}^{G} \Bigg( \min \Bigg( \frac{\pi_{\theta}(o_{i})}{\pi_{\theta_{\text{old}}}(o_{i})} A_{i}, \text{clip}\Big( \frac{\pi_{\theta}(o_{i})}{\pi_{\theta_{\text{old}}}(o_{i})}, 1-\varepsilon, 1+\varepsilon \Big) A_{i} \Bigg) - \beta \mathbb{D}_{KL}\left(\pi_{\theta} \, \| \, \pi_{\text{ref}}\right) \Bigg) \Bigg],
  \end{aligned}
  \label{eq:grpo}
  \end{equation}

where the KL divergence term
$$
\mathbb{D}_{KL} (\pi_\theta || \pi_{ref}) = \frac{\pi_{ref} (o_i|q)} {\pi_\theta(o_i|q)} - \text{log} \frac{\pi_{ref} (o_i|q)} {\pi_\theta(o_i|q)} -1
$$

acts as a regularization term to control divergence from a reference policy $\pi_{ref}$, ensuring stability in training. The hyperparameters $\epsilon$ and $\beta$ regulate the clipping range and the penalty on KL divergence, respectively.

Rather than relying on a learned value function, GRPO computes the advantage function based purely on the observed rewards within a generated group:
$$
A_{i} = \frac{r_{i} - \mathrm{mean}(\{r_1, r_2, \cdots, r_G\})}{\mathrm{std}(\{r_1, r_2, \cdots, r_G\})}
$$

This formulation normalizes rewards within the group, effectively establishing a dynamic “score line”, which serves as a baseline for determining whether a response is better or worse than the group average. By avoiding an explicit value function, GRPO significantly reduces computational overhead and simplifies policy optimization, making it more efficient than PPO.  

Despite its computational efficiency, GRPO presents certain trade-offs when compared to PPO. Xie et al. \cite{xie2025logicrlunleashingllmreasoning} demonstrated that GRPO reduces per-step training time by 60\% compared to PPO, making it particularly attractive for large-scale reinforcement learning applications. Hu \cite{hu2025reinforce++} identified that GRPO is more susceptible to length hacking—a known issue in reinforcement learning where models generate longer responses to artificially increase reward accumulation. PPO’s reliance on an explicit value function appears to mitigate this issue more effectively.

While GRPO accelerates training, it underperforms PPO in accuracy, as the lack of a value function may lead to suboptimal reward estimation, especially in complex tasks requiring precise numerical reasoning.

\subsection{Benchmarks}

To assess the impact of reinforcement learning (RL) training, we systematically compare the performance of our newly optimized model against the base model across six widely recognized benchmarks. These benchmarks have been extensively used to evaluate large language models (LLMs), with existing results available on HuggingFace \cite{open_llm_leaderboard}. The selected benchmarks are: Grade School Math 8K (GSM8k)~\cite{cobbe2021gsm8k}, Instruction-Following Eval (IFEval) \cite{zhou2023instructionfollowingevaluationlargelanguage}, BIG-Bench Hard (BBH) \cite{suzgun2022challenging}, Mathematics Aptitude Test of Heuristics (MATH) \cite{hendrycks2021measuringmathematicalproblemsolving}, 
A More Robust and Challenging Multi-Task Language Understanding Benchmark (MMLU-Pro) \cite{wang2024mmlupro}. 
These benchmarks collectively cover a diverse range of linguistic and cognitive challenges, including instruction following, multi-step reasoning, mathematical problem-solving, expert-level Q\&A, and complex knowledge synthesis. Below, we provide a detailed overview of each benchmark and its significance.

\paragraph{Grade School Math 8K (GSM8K)}

GSM8K is a high-quality dataset designed to benchmark mathematical reasoning in language models. It consists of 8,500 carefully curated grade-school-level math word problems, requiring multi-step reasoning and arithmetic operations. GSM8K is widely used to evaluate a model’s ability to perform complex problem-solving, as it emphasizes logical deduction rather than mere pattern recognition. Due to its structured, human-verified solutions, it has become a gold standard for assessing the reasoning capabilities of large language models, with many studies focusing on improving performance through techniques like chain-of-thought prompting and reinforcement learning.

\paragraph{Instruction-Following Eval (IFEval)}

IFEval is designed to assess an LLM’s instruction-following capabilities through a structured set of verifiable instructions. The dataset consists of 541 prompts, each containing one or more atomic instructions that are objectively testable. These instructions span 25 categories, each with multiple syntactic and parametric variants, ensuring robustness against superficial prompt variations. The evaluation is deterministic: correctness is automatically verified using rule-based programs, providing an unambiguous measure of an LLM’s ability to follow instructions precisely.

\paragraph{BIG-Bench Hard (BBH)}

BBH is a subset of the Beyond the Imitation Game Benchmark (BIG-Bench) \cite{srivastava2022beyond}, a comprehensive suite designed to assess LLMs across reasoning, mathematical problem-solving, commonsense knowledge, coding, and creative tasks. BBH specifically consists of 23 high-difficulty tasks that require advanced multi-step reasoning. These tasks are constructed to challenge models beyond surface-level memorization, making BBH an important test for evaluating the depth of reasoning capabilities in LLMs.

\paragraph{Mathematics Aptitude Test of Heuristics (MATH)}
The MATH benchmark is a dataset of 12,500 problems sourced from leading U.S. mathematics competitions. Unlike standard high-school-level mathematical problems, MATH emphasizes advanced problem-solving techniques in algebra, calculus, geometry, and statistics. Many of these problems require mathematical heuristics rather than formulaic computation, making the benchmark a critical test for evaluating an LLM’s ability to apply reasoning strategies beyond rote calculations.

\paragraph{A More Robust and Challenging Multi-Task Language Understanding Benchmark (MMLU-Pro)}
MMLU-Pro is an enhanced version of the Massive Multitask Language Understanding (MMLU) benchmark \cite{hendrycks2021measuringmassivemultitasklanguage}. Unlike the original MMLU, which is heavily knowledge-driven, MMLU-Pro shifts the focus toward complex reasoning by introducing: More challenging, reasoning-intensive questions, reducing reliance on factual recall. An expanded answer choice set (from 4 options to 10), making random guessing less effective. A refined dataset, removing trivial and noisy questions to enhance benchmark reliability.

\subsection{Training Setup}

We fine-tune the Qwen2.5-3B-Instruct language model \cite{qwen2.5}, a 3-billion parameter instruction-tuned model. The training process is conducted on the Countdown Game dataset \cite{CountdownTasks3to4}, which provides structured arithmetic challenges.

To implement reinforcement learning, we utilize HuggingFace's Transformer Reinforcement Learning (TRL) package \cite{vonwerra2022trl}, which includes a built-in implementation of Group Relative Policy Optimization (GRPO). The training parameters are shown in Table~\ref{tab:trainging_parameters}.

\begin{table}[h]
\caption{Training hyperparameters for RL fine-tuning}
\label{tab:trainging_parameters}
    \centering
    \begin{tabular}{|l|c|}
        \hline
        \textbf{Hyperparameter} & \textbf{Value} \\  
        \hline
        Total training steps & 850 \\  
        \hline
        Batch size & 2 \\  
        \hline
        Learning rate & \( 1.0 \times 10^{-6} \) \\  
        \hline
        GRPO reward calculation & Based on two generated samples per step \\  
        \hline
        KL regularization coefficient (\(\beta\) in Eq. \ref{eq:grpo}) & 0.04 \\  
        \hline
    \end{tabular}
    
    \label{tab:training_hyperparams}
\end{table}

\section{Results and Discussions}

\subsection{Rule Violation in Early Training Stage}

During the initial training phase, the model begins to structure its reasoning using <think>...</think> tags but fails to follow the correct format. Specifically, it introduces multiple <think>...</think> segments, which violates the expected format.

\begin{tcolorbox}[colback=blue!5, colframe=blue!75!black, title=RRule Violation in Early Training Stage]
<think>\\
To get close to 76 with the given numbers and operations, I'll try combinations of the numbers 36, 9, and 49.\\
Let's start with 49. I know that \textcolor{red}{49 - 9 - 9 = 31}, which is close to 76.\\
But still not quite there, so let's add some more numbers.\\
\textcolor{red}{</think>\\
<think>\\
}\\
Considering we have only 76 and our sum so far is 31, let's try 36 + (49 - 9 - 9).\\
</think>\\
<answer> (36 + 49) - 9 - 9 = 76 </answer>\\
\end{tcolorbox}

In this example, the model incorrectly segments its reasoning into two separate <think>...</think> blocks instead of maintaining a single coherent reasoning section. Additionally, the math calculation is incorrect, possibly due to the incapacity of the 3B parameter base model.  

\subsection{Early Sign of Human-like Thinking}\label{earlysignhumanlikethinking}

After 80 training steps, the model successfully adopts the correct response structure, encapsulating its reasoning process within the <think> and <answer> tags. However, instead of demonstrating logical, step-by-step reasoning, it follows a brute-force, trial-and-error approach, systematically testing different arithmetic combinations until it arrives at the correct answer. Notably, at this stage, the model only utilizes addition and subtraction, without incorporating multiplication or division.

This behavior indicates that rather than employing structured problem-solving strategies, the model relies on exhaustive enumeration of possible operations. While this method eventually produces the correct result, it is inefficient and lacks human-like problem-solving heuristics. Further training refinements are needed to encourage the model to develop structured reasoning instead of blindly iterating through possible solutions.

Despite its reliance on trial and error, the model does exhibit a verification-based thinking process by systematically evaluating its computations and comparing the results against the target value (43). This is evident in its responses such as "too high" and "too high again", which suggest an ability to assess and categorize outcomes. However, it lacks true reflection or strategic adaptation. Rather than analyzing incorrect attempts to refine its approach—such as recognizing numerical patterns or adjusting calculations based on previous errors—the model continues testing different combinations until it stumbles upon a valid equation.

\begin{tcolorbox}[colback=blue!5, colframe=blue!75!black, title=Correct Response Structure and Verification-based Thinking]
<think> I need to create an equation that uses the numbers 73, 77, and 39 exactly once to equal 43. Let's try different combinations of operations.\\
- First, let's see if adding or subtracting these numbers gives us 43.\\
- 73 + 77 + 39 = 189 \textcolor{blue}{(too high)}\\
- 73 + 77 - 39 = 111 \textcolor{blue}{(still too high)}\\
- 73 - 77 + 39 = 35 \textcolor{blue}{(too low)}\\
- 39 + 73 + 77 = 189 \textcolor{blue}{(too high again)}\\
- 39 + 77 - 73 = 43 \textcolor{blue}{(this looks promising) }\\
So, the equation works out to be 39 + 77 - 73 = 43.\\
</think>\\
<answer> (39 + 77) - 73 </answer>\\
\end{tcolorbox}

\subsection{Human-like Thinking}

After 300 steps, the model is able to engage in an extensive step-by-step exploration of possible solutions. In the successful example shown below, we can see that it consistently evaluates expressions and adjusts calculations when results do not match the target value. It showcases trial-and-error thinking, similar to how a human might struggle and persist through different approaches. It also shows backtracking when an incorrect equation is identified.

\begin{tcolorbox}[colback=blue!5, colframe=blue!75!black, title=Human-like Thinking]
<think> Firstly, I need to create an equation using the numbers 39, 72, and 95 that is equal to 62. Let's first consider how we can manipulate the numbers. Notice that 95 is the largest and 39 is a relatively small number. One way to get the result close to 62 would be to subtract a large number and then use the remaining numbers to get the final result. Let's try using subtraction first. If I subtract 72 from 95, I get 23. Now I need to manipulate 39 and 23 to reach 62. Adding 39 and 23 gives me 62, which is exactly what we need! </think>  <answer> (95 - 72) + 39 </answer>
\end{tcolorbox}

This example demonstrates the model's systematic exploration and pattern recognition. It correctly identifies the largest number (95) and the smallest (39) and considers how they can be manipulated. It tests a reasonable operation (95-72=23) and then checks if the remaining numbers can be combined to reach 62. This is an efficient and direct approach; it does not waste time testing multiplication or division. The model then verifies the results to ensure the correctness before finalizing the equation. The logic is clear and each step follows from the last. The model also shows reflection when it evaluates different ways of manipulating the numbers.

\subsection{The Aha Moment}

In the R1 report~\cite{guo2025deepseek}, the term "Aha moment" primarily describes the model’s sudden acquisition of complex reasoning behaviors. This phenomenon is often associated with emergent self-correction capabilities and is sometimes signaled by phrases like "Wait" or "Oops", which mimic human-like expressions of reflection. In the R1 report, the team suggested such linguistic cues indicate the model is engaging in real-time reassessment of its thought process, either revising an error or considering an alternative approach to solving the problem.

Our study indicates that responses containing such phrases often reflect an internal debugging process, where the model identifies inconsistencies in its reasoning and attempts to correct them. However, the Aha moment does not necessarily guarantee a correct final answer. Instead, it suggests that the model is exhibiting a structured cognitive process similar to human problem-solving through different strategies, detecting potential miscalculations, and refining its approach.

The following example highlights an Aha moment in action. Here, the model initially follows a straightforward mathematical strategy but soon realizes a flaw in its approach. It then self-corrects by reconsidering the numbers involved, demonstrating an essential aspect of higher-order reasoning. Even though the final answer remains incorrect, the underlying thought process exhibits a structured, step-by-step methodology that is characteristic of intelligent problem-solving.

\begin{tcolorbox}[colback=blue!5, colframe=blue!75!black, title=Aha Moment]\label{ahamoment} <think> First, I can see that none of the numbers are divisible by 3 directly. So, I'll try to break this down using the given operations. </think>\\ <think> One approach could be to use the 84 and 16 to get a number close to 36 with a subtraction operation. </think>\\ <think> I can try subtracting 24 from 84, which gives us 84 - 24 = 60. This is not exactly 36, but it's close. </think>\\ <think> Now, I need to get from 60 to 36. Let's use the 24 we worked with before, and we subtract 24 from 63 (from the list) to get 39. Then subtract 16 to get 18. </think>\\ <think> So, I need a way to get 18 from the remaining numbers. We can use 16 itself, since it's one of the given numbers, and we need to get to 18 from it. </think>\\ <think> I can add 2 to 16 to get 18, so our equation looks like: (84 - 24) - (63 - 61). \textcolor{blue}{But wait, we need 61, not 63.} </think>\\ <think> I realize I need to subtract 2 more from 63 to get 61. So, the final equation is: (84 - 24) - (63 - 2). Let's calculate this. </think>\ <answer> (84 - 24) - (63 - 2) = 36 </answer> \end{tcolorbox}

This instance highlights key characteristics of reasoning development in AI models. The model does not follow a rigid, predefined path but instead explores different possibilities, adjusting its approach as it detects inconsistencies. The phrase "But wait, we need 61, not 63" signifies an active recognition of a prior mistake, similar to how humans verbalize thought processes during problem-solving.
While the model demonstrates a reasoning breakthrough, the final solution still contains an error, indicating that self-reflection alone does not guarantee correctness—it must be paired with a robust verification mechanism. Further analysis into the characteristics of these moments – when they occur, what triggers them, and why they sometimes fail – could provide deeper insights.

\subsection{Response Length vs Reasoning Capability }

Figure~\ref{fig:completion_length} shows the completion length, which fluctuates over time, dropping to its lowest point at 400 steps before increasing and decreasing again. This observation is based on a limited dataset from a simulation of only 850 steps, so the length may eventually rise further if the simulation is extended. Notably, this behavior contrasts with the findings in \cite{guo2025deepseek}, where an increase in length was observed. A closer look at this discrepancy reveals that the model initially relies on a brute-force approach. During the early steps, it explores various combinations in an attempt to find the correct answer. However, as training progresses, the model gradually shifts from this brute-force method to a more reasoning-based approach~\ref{ahamoment}, reflecting its growing ability to solve problems in a more structured, logical way.

\begin{figure}[H]
  \centering
  \includegraphics[width=0.6\textwidth]{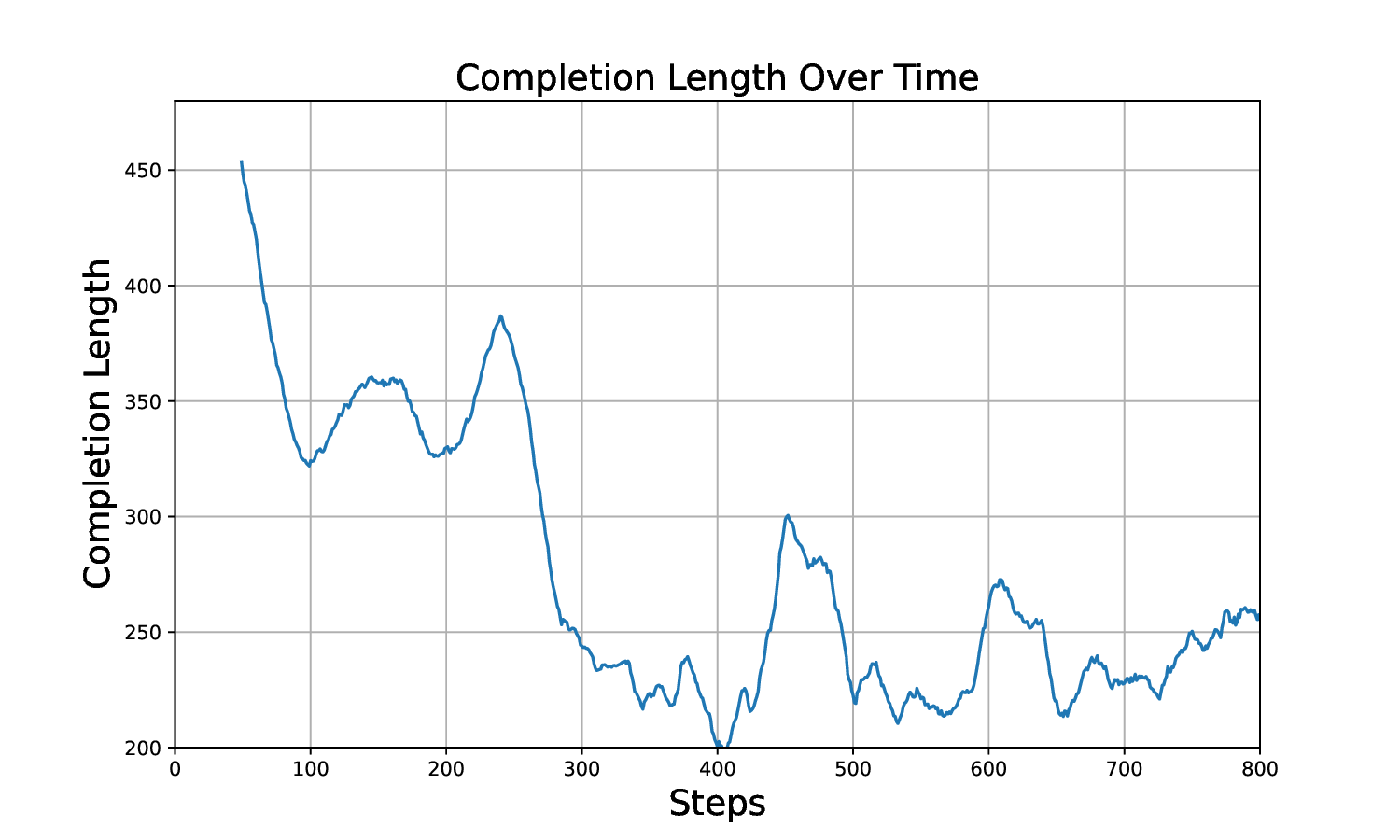}  
  \caption{Change of completion length with training steps}
  \label{fig:completion_length}
\end{figure}

We also observed that, in some instances, the model provides the correct answer without fully revealing its reasoning process~\ref{short_response}. This behavior raises important questions about the model’s decision-making. One possible explanation for this could be that the model is capable of reaching conclusions through internal representations that do not always translate into explicit intermediate steps. In some cases, the model may identify patterns or shortcuts during training that allow it to generate the correct output without the need for detailed reasoning. While this might make the model more efficient in certain situations, it could also limit our understanding of how the model arrives at its answers. The lack of visible reasoning might pose challenges in ensuring the model's reliability, as users would have difficulty verifying the correctness of its decisions. Furthermore, the absence of a clear thought process makes it harder to debug or fine-tune the model, as it would be difficult to pinpoint where the reasoning might go wrong. 

\begin{tcolorbox}[colback=blue!5, colframe=blue!75!black, title=Aha Moment]\label{short_response} <think> First, I should consider how I can use each number to reach 61. With the given numbers 59, 42, and 44, I need to use each exactly once. Let's look for a way to subtract or add numbers to get the difference to 61. Probably multiplication or division won't get me much closer - it's more likely a combination of additions or subtractions will be helpful. </think>
<answer>(59 - 42) + 44</answer>
\end{tcolorbox}

\subsection{Performance of the Trained Model}

Our base model is initially trained on the Countdown Game, where we observe that it develops certain human-like reasoning patterns in its responses. To assess whether these acquired reasoning capabilities generalize beyond the training domain, we conduct a systematic evaluation on six diverse benchmarks spanning different problem domains.  

For performance evaluation, we utilize the Language Model Evaluation Harness~\cite{eval-harness}, a widely adopted framework for benchmarking language models across multiple tasks. We test both the base model and the trained model under standardized conditions:  

\begin{itemize}
    \item  Base Model Evaluation: We use two prompting strategies—(1) the **default prompt** from the evaluation harness and (2) the R1 prompt~\ref{r1_prompt}.  

    \item Trained Model Evaluation: We employ a prompt structure similar to that used in the R1 report~\cite{guo2025deepseek}, ensuring consistency in comparison.  

\end{itemize}

All results are reported under a zero-shot setting unless explicitly stated otherwise, ensuring that performance is measured without any task-specific fine-tuning. This approach allows us to determine whether the model’s reasoning skills, acquired from Countdown Game training, can enhance its performance across a broader range of challenges.

\begin{tcolorbox}[colback=blue!5, colframe=blue!75!black, title=R1 Prompt]\label{r1_prompt} "You are a helpful assistant. You first think about the reasoning process in the mind and then provide the user with the answer. Problem: {{question}} Show your work in <think> </think> tags. Your response should end with \"The final answer is [answer]\" where [answer] is the response to the problem.  Think step by step inside <think> tags. Let me solve this step by <think>"
\end{tcolorbox}

\paragraph{GSM8K}

Table~\ref{tab:GSM8K} presents the model performance on the GSM8K benchmark, evaluated using two metrics: Flexible-Extract and Strict-Match. Flexible-Extract allows some variation in response formatting while assessing correctness, whereas Strict-Match demands an exact match with the reference answer, making it a more rigorous metric.

\begin{table}[H]
 \caption{Benchmark GSM8K}
 \label{tab:GSM8K}
  \centering
  \begin{tabular}{lll}
    \toprule
     Model & Flexible-Extract & Strict-Match                   \\
    \midrule
    Base Model + CoT Llama & 66.3\%  &  29.3\%     \\
    \midrule
    Base Model + R1 Prompt & 60.9\%  &  55.0\%     \\
    \midrule
    Trained Model  + R1 Prompt & 69.7\%  &  64.8\%     \\
    \bottomrule
  \end{tabular}
\end{table}

The Base Model with Chain-of-Thought (CoT) Zero-Shot achieves 66.3\% on Flexible-Extract, but only 29.3\% on Strict-Match. After checking the answers, we found that the output format does not consistently follow a specific pattern. As a result, the LLM Harness struggles to determine the correctness of the response, even though a human evaluator would consider the answer accurate. 

Adding the R1 Prompt to the Base Model significantly boosts Strict-Match accuracy (55.0\%), showing that the prompt improves both reasoning quality and consistency in answer formatting. The Trained Model with R1 Prompt outperforms all base model configurations, achieving 69.7\% on Flexible-Extract and 64.8\% on Strict-Match. This demonstrates that reinforcement learning (or fine-tuning) enhances both reasoning depth and answer consistency, leading to more structured and accurate responses.

We experimented with few-shot learning for the trained model; however, its performance was worse than in the zero-shot setting. This result aligns with the findings reported in \cite{guo2025deepseek}, which also observed that few-shot learning underperformed relative to zero-shot performance. The reason is unknown.  Further investigation into the model's behavior with different few-shot configurations may help clarify this issue.

We compare the answers from the base model and from the trained model to highlight the differences. The question is \textit{John runs 60 miles a week. He runs 3 days a week.  He runs 3 hours the first day and half as much the other two days he runs.  How fast does he run?}

The model begins with an incorrect assumption about the total number of hours that John runs. It computes 3 × 6 = 18 hours, which is a misunderstanding of the structure of the problem. 
Next, the model incorrectly states that John runs 3 hours on the third day instead of correctly recognizing that he runs half of the first day's time (i.e., 1.5 hours). This introduces another error, leading to a total of 7.5 hours per week.

\begin{tcolorbox}[colback=blue!5, colframe=blue!75!black, title=Base model answer to one GSM8K problem ]\label{base_model_gsm8k} \textcolor{red}{He runs 18 total hours because 3 x 6 = 18}. He runs 3 hours the first day then 3 / 2 = 1. 5 hours the second day. \textcolor{red}{He runs 3 hours the third day}. He runs a total of 3 + 1. 5 + 3 = 7. 5 hours per week. So his speed is 60 / 7. 5 = 8 miles per hour. Therefore, the answer is 8.
\end{tcolorbox}

The trained model begins by correctly calculating the running hours for each day. It uses the given information that John runs 3 hours on the first day and half that amount (1.5 hours) on the other two days. These values are accurately calculated.
The model correctly adds the total weekly running hours: 3 + 1.5 + 1.5 = 6 hours. The trained model correctly interprets the question, using the information to break down the problem step by step. The Trained Model demonstrates a clear understanding of the problem and successfully breaks it down into manageable parts, ensuring accurate results.

\begin{tcolorbox}[colback=blue!5, colframe=blue!75!black, title=Trained model answer to one GSM8K problem ]\label{r1_model_gsm8k} <think> First, let's determine how many hours John runs on each of the two days he runs half as much as the first day. We know that John runs a total of 3 hours on the first day. On the other two days, he runs half as much, so each of these days he runs 3 / 2 = 1.5 hours. \\

Next, we calculate the total running hours each day: \\
- First day: 3 hours \\
- Second day: 1.5 hours \\
- Third day: 1.5 hours \\

Summing these up gives the total weekly running hours:
3 + 1.5 + 1.5 = 6 hours

Now we know John runs a total of 60 miles in a week and runs for a total of 6 hours. To find his average speed, we use the formula:

Speed = Total Distance/Total Time

Substituting the given values:

Speed= 60 miles/6 hours

Speed = 10 miles per hour

</think>The final answer is 10.

\end{tcolorbox}

\paragraph{Math}

Table~\ref{table:minerva_math} presents the performance on the Math benchmark, evaluated using two metrics: Extract Match and Math Verify. We use \textit{minerva math} from the LLM Harness. We changed the setting to run a zero-shot evaluation. The evaluation used 5000 problems instead of the 12,500 ones.  We compare the math verify result instead of the exact match because the exact match results are zero for both models. This indicates that neither model consistently produces answers that exactly match the reference solutions in the expected format. The lack of improvement even after training suggests that format inconsistencies or structural variations in responses remain a challenge.

The Math Verify metric, which assesses whether the final numerical answer is mathematically correct regardless of formatting, shows a clear improvement after training.

\begin{table}[ht]
\caption{Comparison of Model Performance  on the MATH Benchmark}
\centering
\begin{tabular}{|l|c|c|}
\hline
\textbf{Tasks} & \textbf{Base Model (math\_verify)} & \textbf{Trained Model (math\_verify)} \\
\hline
 Math Algebra                 & 0.1837 & 0.4027 \\
 Math Counting and Probability & 0.0970 & 0.2489 \\
 Math Geometry                 & 0.0438 & 0.1983 \\
 Math Intermediate Algebra     & 0.0520 & 0.0997 \\
 Math Number Theory            & 0.1019 & 0.1926 \\
 Math Prealgebra               & 0.2767 & 0.5109 \\
 Math Precalculus              & 0.0403 & 0.0769 \\
\hline
\textbf{Group Result}  & 0.1300 & 0.2744 \\
\hline
\end{tabular}

\label{table:minerva_math}
\end{table}

The Base Model achieves 13\%, while the Trained Model with R1 Prompt improves to 27\%, representing a relative increase of more than 100\%. This suggests that training enhances the model's reasoning ability, allowing it to solve more problems correctly, even if its output format does not strictly adhere to the reference answer format.

We compare answers from both models. The problem is \textit{What is the smallest whole number $b$ such that 47 can be expressed in base $b$ using only two digits?}. The base model has incorrect reasoning at Step 2 when it said \textcolor{red}{We need $47 = ab + c$ where $ a \geq b-1 $ and $ 0 \leq c < b $.}. It shold be $a\le b-1 $.  The trained model recognize the if the number 47 can be represented using only two digits, then there should be a $k$ so that $b^k \ge 47 > b^{k-1} $. The better way is to set $k=2$ because it uses only two digits. 

\begin{tcolorbox}[colback=blue!5, colframe=blue!75!black, title=Base model answer to one MATH Number Theory problem]\label{base_model_math}
  To determine the smallest whole number \( b \) such that the number 47 can be expressed in base \( b \) using only two digits, we need to follow these steps: 
  
  1. **Understand the representation of a number in base \( b \) with two digits:**
  
  - If a number \( N \) can be expressed with two digits in base \( b \), it can be written as:
  \[
  N = ab + c
  \]
  where \( a \) and \( c \) are digits in base \( b \). For a two-digit number in base \( b \), the possible values for \( a \) (the leading digit) are \( b-1 \) or higher, and \( c \) (the trailing digit) ranges from 0 to \( b-1 \).
  
  2. **Express \( 47 \) as a two-digit number in base \( b \)**
  
  - \textcolor{red}{We need \( 47 = ab + c \) where \( a \geq b-1 \) and \( 0 \leq c < b \).}
  
  3. **Set up the inequality for the digits:**
  
  - Since \( ab + c = 47 \) and \( a \geq b-1 \), we can rearrange to solve for \( c \):
  \[
  c = 47 - ab
  \]
  - For \( c \) to be a valid digit in base \( b \):
  \[
  0 \leq 47 - ab < b
  \]
  
  4. **Determine the smallest \( b \):**
  
  We need to find the smallest \( b \) such that there exists an integer \( a \) where \( b-1 \leq a < b \) and \( 0 \leq 47 - ab < b \).
  
  Let's test small values of \( b \):
  
  - For \( b = 5 \):
  \[
  47 - ab < 5 \Rightarrow 47 - a \cdot 5 < 5 \Rightarrow 42 < 5a \Rightarrow a > 8.4
  \]
  The smallest integer greater than 8.4 is 9.
  - So, \( a = 9 \) and \( 47 - 9 \cdot 5 = 47 - 45 = 2 \).
  - Here, \( ab + c = 9 \cdot 5 + 2 = 45 + 2 = 47 \). We need 2 digits to represent 47 in base 5.
  
  - For \( b = 6 \):
  \[
  47 - ab < 6 \Rightarrow 47 - a \cdot 6 < 6 \Rightarrow 41 < 6a \Rightarrow a > 6.83
  \]
  The smallest integer greater than 6.83 is 7.
  - So, \( a = 7 \) and \( 47 - 7 \cdot 6 = 47 - 42 = 5 \).
  - Here, \( ab + c = 7 \cdot 6 + 5 = 42 + 5 = 47 \). We need 2 digits to represent 47 in base 6.
  
  Since we are looking for the smallest \( b \) satisfying the condition, we see that \( b = 5 \) works while \( b = 6 \) also works. Therefore, the smallest whole number \( b \) such that 47 can be expressed in base \( b \) using only two digits is \( 5 \).
  
  \[
  \boxed{\textcolor{red}{5}}
  \]
  \end{tcolorbox}

\begin{tcolorbox}[colback=blue!5, colframe=blue!75!black, title=Trained model answer to one MATH Number Theory problem ]\label{trained_model_math} 

We need to find the smallest base b such that the decimal number 47 can be represented using only two digits in base b. A number n can be represented as abcd... in base b if and only if $b^k >= 47 > b^{k-1}$ for some k, where a, b, c, d... are the digits of the number in base b. This means we need to find the smallest b such that $b^2 > 47$. Let's check the squares of successive integers starting from 1 until we find the smallest b satisfying this condition. $1^2 = 1$ $2^2 = 4$ $3^2 = 9$ $4^2 = 16$ $5^2 = 25$ $6^2 = 36$ $7^2 = 49$ 

We see that $6^2 = 36 < 47$ and $7^2 = 49 > 47$. Therefore, the smallest base b is 7.</think>

The smallest whole number b such that 47 can be expressed in base b using only two digits is 7. \textcolor{blue}{So, the answer is 7}.

\end{tcolorbox}

\paragraph{BBH}

Table~\ref{table:bbh_model_performance} presents the Extract Match accuracy of different models on the BBH benchmark, which evaluates the model’s ability to generate correct answers in a strictly matched format. he Base Model achieves an Extract Match score of 37.6\%, indicating its baseline capability to generate correct, well-formatted answers. When applying the R1 Prompt to the Base Model, performance slightly decreases to 36.5\%. The Trained Model with R1 Prompt achieves 44.6\%, a notable improvement of 7\% points over the Base Model.

\begin{table}[ht]
\caption{Comparison of Model Performance across Different Tasks in BBH}
\centering
\begin{tabular}{|l|c|c|c|}
\hline
\textbf{Tasks} & \textbf{Base Model} & \textbf{Base Model + R1 Prompt} & \textbf{Trained Model} \\
\hline
\textbf{Group Result}                               & 0.375      & 0.365      & 0.4406 \\
\hline
Boolean Expressions                                 & 0.8280 & 0.7840 & 0.9040 \\
Causal Judgement                                    & 0.4439 & 0.4492 & 0.4920 \\
Date Understanding                                  & 0.3560 & 0.2480 & 0.4920 \\
Disambiguation QA                                   & 0.4200 & 0.3800 & 0.4960 \\
Dyck Languages                                      & 0.0080 & 0.0240 & 0.0040 \\
Formal Fallacies                                    & 0.4720 & 0.3440 & 0.5040 \\
Geometric Shapes                                    & 0.0800 & 0.0520 & 0.1280 \\
Hyperbaton                                          & 0.5520 & 0.5960 & 0.5560 \\
Logical Deduction (Five Objects)                    & 0.3440 & 0.3240 & 0.3360 \\
Logical Deduction (Seven Objects)                   & 0.2680 & 0.3160 & 0.3440 \\
Logical Deduction (Three Objects)                   & 0.3080 & 0.2320 & 0.6200 \\
Movie Recommendation                                & 0.4760 & 0.3520 & 0.3760 \\
Multistep Arithmetic (Two)                          & 0.1760 & 0.2000 & 0.6000 \\
Navigate                                            & 0.7480 & 0.7600 & 0.7200 \\
Object Counting                                     & 0.0520 & 0.0400 & 0.1200 \\
Penguins in a Table                                 & 0.6712 & 0.7055 & 0.8014 \\
Reasoning about Colored Objects                     & 0.5000 & 0.5240 & 0.6840 \\
Ruin Names                                          & 0.3560 & 0.3480 & 0.2760 \\
Salient Translation Error Detection                 & 0.1920 & 0.2120 & 0.4040 \\
Snarks                                              & 0.5000 & 0.5562 & 0.5955 \\
Sports Understanding                                & 0.5560 & 0.5400 & 0.4920 \\
Temporal Sequences                                  & 0.3040 & 0.3360 & 0.2360 \\
Tracking Shuffled Objects (Five Objects)            & 0.3720 & 0.3360 & 0.2680 \\
Tracking Shuffled Objects (Seven Objects)           & 0.2200 & 0.2320 & 0.1840 \\
Tracking Shuffled Objects (Three Objects)           & 0.3760 & 0.4000 & 0.6320 \\
Web of Lies                                         & 0.6560 & 0.5400 & 0.7400 \\
Word Sorting                                        & 0.1000 & 0.1240 & 0.1000 \\
\hline
\end{tabular}
\label{table:bbh_model_performance}
\end{table}

The most significant improvements are from Date Understanding,
Disambiguation QA, Logical Deduction (Three Objects), Reasoning about Colored Objects, and  Tracking Shuffled Objects (Three Objects).  We see less improvement in tasks like Boolean Expressions or Object Counting, which are more mechanical and involve straightforward processing of information.

\paragraph{MMLU Pro}

Table~\ref{table:mmlu_pro_comparison} presents the test results on the  MMLU Pro benchmark, which evaluates a model’s ability to generate correct answers across diverse professional-level knowledge domains.

The Base Model achieves an Extract Match score of 16.3\%, indicating its baseline ability to produce correct responses that match the reference format. This relatively low score suggests that the Base Model struggles with either formatting consistency or accurate recall of domain-specific knowledge. The Trained Model with R1 Prompt significantly improves to 22.4\%, an absolute increase of 6.2\% over the Base Model and a 9.1\% improvement over the Base Model + R1 Prompt.

The improved performance of the trained model is evident across multiple disciplines, particularly in Psychology (+68.5\%), Biology (+60.5\%), and Mathematics (+38.6\%), indicating a broad enhancement in both factual recall and complex reasoning capabilities. This suggests that fine-tuning has strengthened the model’s ability to integrate domain-specific knowledge with logical inference. The improvement in Psychology and Biology may stem from enhanced pattern recognition and causal reasoning, essential for understanding behavioral and biological systems.  These results demonstrate that cross-disciplinary advancements in AI models are achievable through reinforcement learning.

\begin{table}[ht]
\centering
\caption{Comparison of Base Model and Trained Model Performance on MMLU Pro Tasks}
\begin{tabular}{|l|c|c|}
\hline
\textbf{Tasks} & \textbf{Base Model} & \textbf{Trained Model} \\
\hline
Biology           & 0.2329 & 0.3738 \\
Business          & 0.2459 & 0.3016 \\
Chemistry         & 0.0574 & 0.1087 \\
Computer Science  & 0.2049 & 0.2537 \\
Economics         & 0.2571 & 0.3128 \\
Engineering       & 0.0485 & 0.0640 \\
Health           & 0.1443 & 0.2347 \\
History          & 0.1549 & 0.1890 \\
Law              & 0.0827 & 0.0954 \\
Math             & 0.2339 & 0.3242 \\
Other            & 0.2348 & 0.2911 \\
Philosophy       & 0.1804 & 0.2044 \\
Physics          & 0.1039 & 0.1586 \\
Psychology       & 0.1942 & 0.3271 \\
\hline
\textbf{Group Result (MMLU Pro)} & 0.1625 & 0.2242 \\
\hline
\end{tabular}

\label{table:mmlu_pro_comparison}
\end{table}

\paragraph{IFEval}

The Base Model performed well with a Loose Accuracy of 66.0\% and a Strict Accuracy of 60.3\%, demonstrating a solid baseline performance without additional fine-tuning or prompt engineering.

For the trained model, the Loose Accuracy decreased slightly to 64.8\%, and the Strict Accuracy showed a minor decrease to 60.1\%. 

\begin{table}[H]
 \caption{Benchmark IFEval}
 \label{tab:ifeval}
  \centering
  \begin{tabular}{lll}
    \toprule
     Model & Loose Accuracy& Strict Accuracy                   \\
    \midrule
    Base Model & 66.0\%  &  60.3\%     \\
    \midrule
    Trained Model  + R1 Prompt  & 64.8\%  &  60.1\%     \\
    \bottomrule
  \end{tabular}
  \label{tab:table}
\end{table}

\section{Conclusion and Future Work}

This work demonstrates the potential of pure RL in enhancing the reasoning capabilities of language models. Our experiments show that the RL-trained model outperforms the base model on four out of five benchmarks, indicating that RL can drive significant improvements in generalization and task performance. While we successfully replicated the "aha moment", we found that these moments did not directly contribute to higher accuracy. Additionally, we discovered reasoning responses became shorter as the model transition from brute-force approaches to a more human-like reasoning approaches. 

During our training process, we observed several issues that impacted the model’s performance and evaluation:

\begin{itemize}
    \item Response Format Violations: At the end of the training period, the model still occationally generates wrong formats like duplicated <answer></answer> blocks. 

\item Model Evaluation Harness Limitations: The evaluation harness we used misjudged the quality of model responses when model responses deviated from the expected format. A more robust evaluation framework, potentially incorporating a stronger model, could better assess and validate the model’s outputs.

\item Performance with Few-Shot Prompts: The trained model did not perform as well when using few-shot prompts, a result consistent with prior findings~\cite{guo2025deepseek}.

\item Reward Function Judgments: The format-reward and answer-reward functions occasionally misjudged the model responses, especially in cases where human evaluators would deem an answer correct, but the reward functions assessed it as incorrect. 

\item GRPO: We need to examine the impact of sample size on the evaluation of the advantage function during training. The number of samples used may significantly affect the performance and stability of the model, and this aspect requires further exploration to optimize training efficacy.

 \end{itemize}

\bibliographystyle{unsrt}
\bibliography{references}

\end{document}